\documentclass{article}
\usepackage{spconf,amsmath,graphicx}
\usepackage{amsfonts,amssymb,commath}
\usepackage{bibspacing}
\title{ABNORMAL CHEST X-RAY IDENTIFICATION WITH GENERATIVE ADVERSARIAL ONE-CLASS CLASSIFIER}
\name{Yuxing Tang$^1$ \qquad Youbao Tang$^1$ \qquad Mei Han$^2$ \qquad Jing Xiao$^3$ \qquad Ronald M. Summers$^1$
}
\address{$^1$Imaging Biomarkers and Computer-Aided Diagnosis Lab, Radiology and Imaging Sciences,\\National Institutes of Health (NIH) Clinical Center, Bethesda, USA;\\
$^2$Ping An Technology, US Research Labs, USA; $^3$Ping An Technology Co., Ltd., Shenzhen, China}
\begin{document}
%
\maketitle
\vspace{-2mm}
\begin{abstract}
Being one of the most common diagnostic imaging tests, chest radiography requires timely reporting of potential findings in the images. In this paper, we propose an end-to-end architecture for abnormal chest X-ray identification using generative adversarial one-class learning. Unlike previous approaches, our method takes only normal chest X-ray images as input. The architecture is composed of three deep neural networks, each of which learned by competing while collaborating among them to model the underlying content structure of the normal chest X-rays. Given a chest X-ray image in the testing phase, if it is normal, the learned architecture can well model and reconstruct the content; if it is abnormal, since the content is unseen in the training phase, the model would perform poorly in its reconstruction. It thus enables distinguishing abnormal chest X-rays from normal ones. Quantitative and qualitative experiments demonstrate the effectiveness and efficiency of our approach, where an AUC of 0.841 is achieved on the challenging NIH Chest X-ray dataset in a one-class learning setting, with the potential in reducing the workload for radiologists. 
\end{abstract}
\vspace{-1mm}
\vspace{-1mm}

\vspace{-1mm}
\section{Introduction}
\label{sec:intro}
\vspace{-1mm}
The chest radiograph (chest X-ray, or CXR) is the most commonly requested radiological examination owing to its effectiveness in the characterization and detection of cardiothoracic and pulmonary abnormalities. It is also widely used in lung cancer prevention and screening. Timely radiologist reporting of every image is desired, but not always possible due to heavy workload. Consequently, an automatic system of CXR abnormality classification would be advantageous, allowing reporting works focusing more on pathology analysis of abnormal CXRs.

Recently, deep learning based approaches have been proposed as the solution to automatic classification and detection of abnormalities in CXRs with promising results~\cite{Wang_CVPR2017, Yates, Tang_MLMI}. A large set of labeled training data are required to build discriminative convolutional neural networks (CNNs) for such a purpose. However, it is not always possible to include or annotate all kinds of abnormalities for large scale training, for the reason that some forms of anomaly are very rare, while on the other hand, normal CXRs are much easier to obtain.

Inspired by one-class classification~\cite{MOYA1996463, sabokrou2018}, which tries to classify data of a specific category among all data by learning from a training set containing only the data of that unique category, in this paper, we present a method that can automatically identify abnormal CXRs by learning only from normal ones: capturing special characteristics of normal CXR collection and figuring out how the unseen abnormal collection differentiates from them. More specifically, we propose an end-to-end generative adversarial one-class learning approach, for normal versus abnormal CXR classification, by training solely from normal CXRs. The proposed architecture, similar to generative adversarial networks (GANs)~\cite{Goodfellow_GAN}, is composed of three main modules: a U-Net autoencoder, a CNN discriminator and an encoder, which compete to learn while collaborating with each other for the target task. The adversarially trained generative model is capable of reconstructing the normal CXRs while performing poorly on reconstructing the abnormal ones, since only the normal CXRs are involved in training and those with various anomalies are unseen by the model. Such reconstruction differentiation enables the proposed model to identify abnormal CXRs.

Previous work ~\cite{Madani18, Salehinejad} adopted GANs to synthesize CXRs in order to augment the training set for classifying abnormalities with less training samples. Although we are reconstructing CXRs, we are not augmenting the training set and we only use CXR samples from a single class to train a one-class classifier. In our work, the reconstructed CXRs are considered as enhancing the inlier (normal) samples and distorting the outlier (abnormal) samples.
\vspace{-1mm}

\begin{figure*}[t!]
  \centering
  \vspace{-2mm}
  \includegraphics[width=0.7\linewidth]{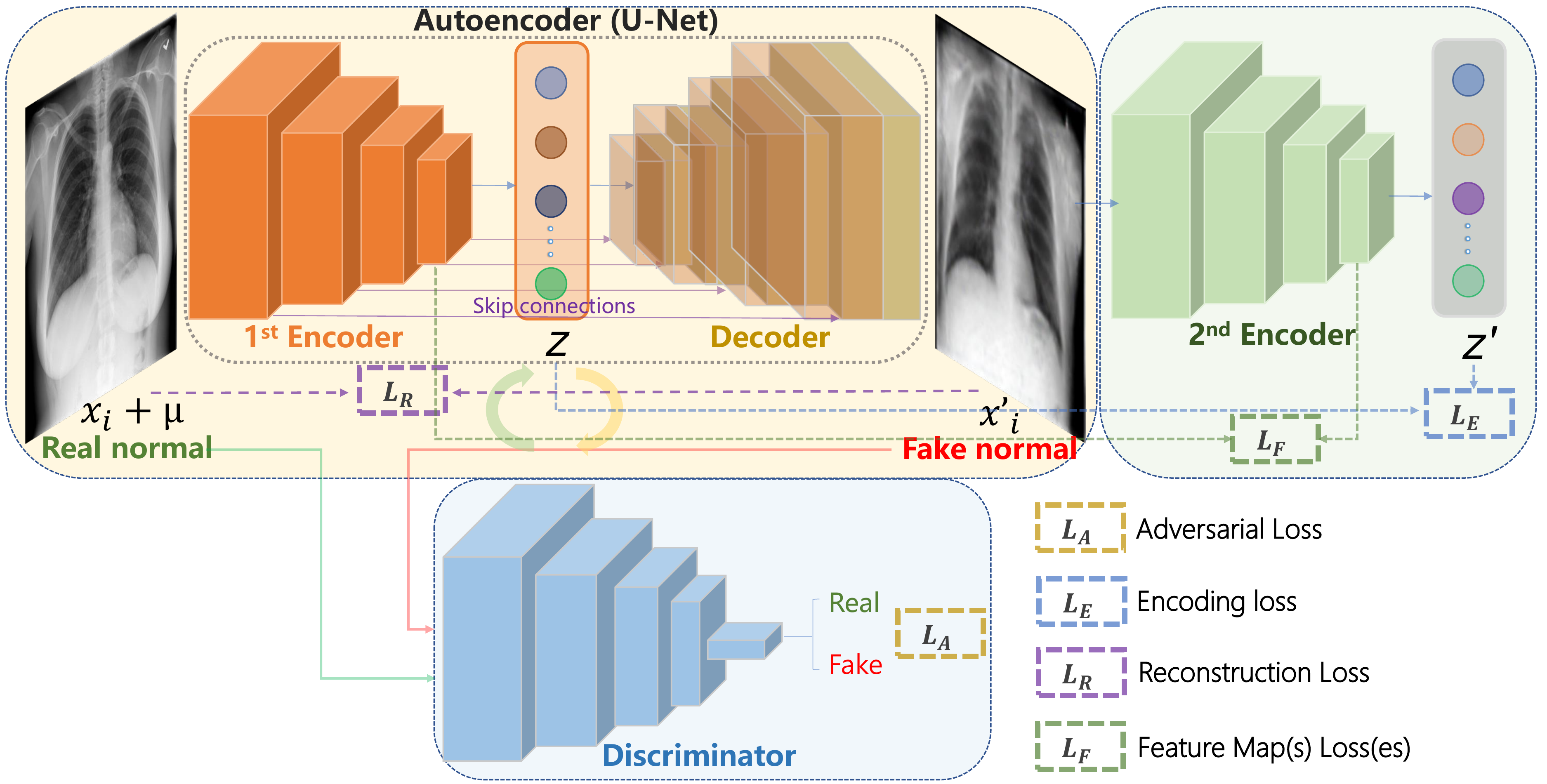}
  \vspace{-1mm}
  \caption{Framework of the proposed deep adversarial one-class learning model for abnormal chest X-ray identification.}
  \vspace{-1mm}
  \label{fig:pipeline}
\end{figure*}

\vspace{-1mm}
\section{Proposed Method}
\label{sec:format}
\vspace{-1mm}
\subsection{Problem Formulation}
\vspace{-1mm}
In our one-class learning scenario, given a training set $\boldsymbol{T} =\{(x_i, y_i), i = 1, \cdots, N\}$, where $N$ is the number of samples in $\boldsymbol{T}$, $x_i$ is a normal CXR image with label $y_i$ ($\forall i$, $y_i=0$). The test set $\boldsymbol{S} =\{(x_j, y_j), j = 1, \cdots, M\}$ contains $M$ CXRs, each of which is labeled as either normal ($y_j = 0$) or abnormal ($y_j = 1$). In the training stage, the goal is to learn a function $f$ from $\boldsymbol{T}$, which models the data distribution of normal CXRs, meanwhile, minimizes the anomaly score $\mathcal{A}(x_i)$ so that $\mathcal{A}((x_i, y_i=0))\rightarrow 0$. In the inference phase, $f$ is expected to output smaller anomaly scores given normal X-ray images and larger scores given abnormal X-rays, such that $\mathcal{A}(x_i, y_i=0) < \mathcal{A}(x_j, y_j=1)$, thus differentiating between normal and abnormal CXRs.

\vspace{-1mm}
\subsection{The Proposed Architecture} 
\vspace{-1mm}
The proposed adversarial one-class learning framework is inspired by the generative adversarial networks (GANs)~\cite{Goodfellow_GAN}.
GAN is formulated as a two-player game, where the generator $\mathbf{G}$ takes a random noise vector in a latent space as input and produces a sample in the data space while the discriminator $\mathbf{D}$ identifies if a certain sample comes from the true data distribution or $\mathbf{G}$. The training procedure is to solve a minimax problem which alternates between training $\mathbf{D}$ and $\mathbf{G}$, such that $\mathbf{G}$ is optimized to generate samples that are not distinguishable by $\mathbf{D}$.

\textbf{Network Architecture:} Three essential modules, namely, a U-Net~\cite{Ronneberger} like autoencoder (generator), a convolutional neural network (discriminator) and an encoder network, together constitute the generative adversarial one-class learning architecture (See Fig.~\ref{fig:pipeline}). The U-Net like \textbf{autoencoder} (denoted as $\mathbf{U}$) first maps an input CXR image $x_i \in \boldsymbol{T}$ with Gaussian noise $\mu$ into a lower-dimensional latent space $z$ using a fully convolutional network ($1^{st}$ encoder  $\mathbf{U}_\mathcal{E}$), which is then inversely mapped back using a deconvolutional network (decoder  $\mathbf{U}_\mathcal{D}$) to generate the reconstructed image $x'_i \in \boldsymbol{T'}$. The U-Net like encoder-decoder with skip connections is adopted to preserve high-resolution features through concatenation in the up-sampling (deconvolution) process, and a CNN \textbf{discriminator} (denoted as $\mathbf{D}$) is looped for adversarial training to produce better and more realistic reconstruction. The first two modules function as a conditional GAN, which are conditioned on the real input image $x_i$. A second \textbf{encoder} $\mathbf{E}$ is padded after the autoencoder, which further encodes the generated fake image into another latent space $z'$, in order to force the consistency between two latent vectors $z$ and $z'$ and corresponding intermediate feature maps from the two encoders.

The intuition behind the proposed framework is that it is able to reconstruct the normal CXRs while performing poorly on reconstructing the abnormal ones, since only the normal chest X-rays are used for training, and the abnormal CXR image contents are unseen. Such differentiation of reconstruction behaviors enables distinguishing abnormal CXRs from normal ones. 

\textbf{Loss Functions:}
The objective of the proposed architecture is to jointly optimize the three modules in an end-to-end manner. To this end, we design four different loss functions:

The \textit{image reconstruction loss} of the autoencoder $\mathcal{L_{R}}$ is formulated as the mean absolute error ($l_1$-norm) of a real input CXR image $x_i$ and its reconstructed image $x'_i$, to measure the similarity between the image pairs: 
\vspace{-1mm}
\begin{equation}
	\mathcal{L_{R}} = \norm{x_i-x'_i}_1, \text{where } 
	x'_i = \mathbf{U}_\mathcal{D}(\mathbf{U}_\mathcal{E}(x_i)).
\vspace{-1mm}
\end{equation}

The \textit{adversarial learning loss} $\mathcal{L_{A}}$ of the conditional GAN is modeled as the binary cross-entropy loss for classification of real CXR $x_i \in \boldsymbol{T}$ and generated fake CXR $x'_i \in \boldsymbol{T'}$:
\vspace{-1mm}
\begin{equation}
\begin{aligned}
\label{eq:gan_loss}
        \mathcal{L_{A}} = \min_{\mathbf{U}} \max_{\mathbf{D}} \mathbb{E}_{x \sim \mathbf{T}} 
         \Big[ \log p(y_i=1 \mid x_i, \mathbf{D}) \Big] +\\ 
         \mathbb{E}_{x'_i \sim \mathbf{T'}}
 		\Big[ \log \big ( 1 - p(y_i=1 \mid x'_i, \mathbf{D}) \big ) \Big], 
 		 x'_i = \mathbf{U}(x_i).
\end{aligned}
\vspace{-1mm}
\end{equation}

The \textit{encoding consistency loss} models the consistency between the two latent space $z$ and $z'$, which is formulated as the mean square error ($l_2$-norm):
\vspace{-1mm}
\begin{equation}
	\mathcal{L_{E}} = \norm{\mathbf{U}_\mathcal{E}(x_i) - \mathbf{E}(x'_i)}_2.
\vspace{-1mm}
\end{equation}

The \textit{feature map consistency loss} measures the overall similarity between intermediate feature maps of the two encoders:
\vspace{-1mm}
\begin{equation}
	\mathcal{L_{F}} = \sum_{l}{\norm{\mathcal{F}_l(\mathbf{U}_\mathcal{E}(x_i)) - \mathcal{F}_l(\mathbf{E}(x'_i))}_2},
\vspace{-4mm}
\end{equation}
where  $\mathcal{F}_l(\cdot)$ is the feature map of the $l^{th}$ layer of the encoder.

The final objective function is: 
\vspace{-1mm}
\begin{equation}
    \mathcal{L} = \lambda_1 \mathcal{L_{R}} + \lambda_2 \mathcal{L_{A}} + \lambda_3 \mathcal{L_{E}} +  \mathcal{L_{F}},
\vspace{-2mm}
\end{equation}
where all $\lambda$s $> 0$ are trade-off hyperparameters that control the relative importance of each of the four terms.

\vspace{-1mm}
\subsection{Inference}
\vspace{-1mm}
In the testing phase, a CXR $x_j$ is passed through the framework and an anomaly score $\mathcal{A}$ is calculated by:

\begin{equation}
\vspace{-1mm}
	\mathcal{A}(x_j) = \lambda_1\norm{x_j-x'_j}_1 +  \lambda_2(1-\mathbf{D}(x_j')) + \lambda_3\norm{z_j-z'_j}_2,
\end{equation}
where $\mathbf{D}(x_j')$ indicates the likelihood that a generated CXR from $x_j$ looks realistic, and $\mathcal{A}(x_j)$ is normalized to $[0,1]$ on all testing samples for binary classification evaluation. Ideally, the anomaly scores of any abnormal samples $\mathcal{A}(x_j, y_j=1)$ should be larger than the scores from normal samples $\mathcal{A}(x_j, y_j=0)$.

\vspace{-1mm}
\section{EXPERIMENTAL RESULTS}
\label{sec:result}
\vspace{-1mm}

\subsection{Dataset and Implementation Details}
\label{sec:data}

We evaluated the proposed framework for normal versus abnormal CXR classification on the NIH Clinical Center Chest X-ray dataset\footnote{https://nihcc.app.box.com/v/ChestXray-NIHCC}~\cite{Wang_CVPR2017}
, which contains 112,120 frontal-view CXR images of 30,805 unique patients. Cardiothoracic and  pulmonary abnormalities include cardiomegaly, lung infiltrate, mass, nodule, pneumonia, pneumothorax, pulmonary atelectasis, consolidation, edema, emphysema, fibrosis, hernia, pleural effusion and thickening. We performed two experiments on this dataset: In the first experiment, we used 4,479 normal (without any abnormal pulmonary or cardiac findings) and no (zero) abnormal CXRs for training, 849 normal and 857 abnormal CXRs for validation, 677 normal and 667 abnormal CXRs for testing. In this experiment, the abnormal CXRs contain at least one of the above 14 findings. In the second experiment, we classified the CXRs with lung opacities (visual signal for pneumonia) from normal CXRs\footnote{https://www.kaggle.com/c/rsna-pneumonia-detection-challenge}. 6,000 normal CXRs were used for one-class training, 1,025 normal CXRs and 1,025 CXRs with lung opacities were used for validation, 1,000 normal and 1,000 CXRs with lung opacities for testing. The training/validation/testing subsets were split by patient ID so there was no patient overlap among these three subsets.

The framework was implemented using the PyTorch library. The U-Net autoencoder consists of a 5-layer CNN encoder and a 5-layer deconvolutional decoder with skip connections (both have batch normalization and leaky ReLU after each layer except for the first layer). The second encoder and discriminator has the similar structure as the encoder in the autoencoder. $4 \times 4$ kernels were used in both down and up sampling, and the latent vector size was 100. The images were resized to $64 \times 64$ pixels, with a batch size of 64. The network was initialized with standard normal distribution and optimized using Adam gradient descent optimizer (momentums $\beta _1$ = 0.5, $\beta _2$ = 0.999) with an initial learning rate of $5e^{-4}$. $\lambda_1$, $\lambda_2$ and $\lambda_3$ were empirically set to 20, 4 and 8 respectively. The training converged in 15 epochs, taking only 5-7 minutes on an NVIDIA TITAN Xp GPU. The inference time for each CXR was 0.48 ms on average. 
 
\subsection{Quantitative and Qualitative Results}

We first quantitatively evaluate the classification performance using the area under the receiver operating characteristic (AUC) score. We conduct ablation studies to the proposed framework. The baseline method is the U-Net autoencoder $\mathbf{U}$ truncated from the proposed model. Then we add a second encoder $\mathbf{E}$ and decoder $\mathbf{D}$ to $\mathbf{U}$ respectively, to compare their performance with the whole network ($\mathbf{U}+\mathbf{D}+\mathbf{E}$). The training and evaluation process is repeated five times to reduce randomness. The normal and abnormal (or lung opacities) CXR classification results in terms of AUCs are shown in Table \ref{tab:results}. The proposed model achieves an average AUC of 0.841 on the testing set for normal and abnormal CXRs with 14 major findings, and an average AUC of 0.802 for normal versus abnormal CXRs with lung opacities, which largely outperforms the baseline U-Net model without adversarial learning. Each of the proposed modules contributes to the final model improvement. The overall result is encouraging given the fact that the dataset contains difficult cases (\textit{e.g.}, mild abnormalities) and only normal CXRs are used in the training stage.

\begin{table}[ht]\fontsize{10pt}{10pt}\selectfont
\renewcommand{\arraystretch}{1.3}
\vspace{-2mm}
\caption{Comparison of classification performance in terms of AUC on the test sets from two experiments. ($\mathbf{U}$: U-Net autoencoder, $\mathbf{E}$: second encoder, $\mathbf{D}$: discriminator.)} 
\vspace{-2mm}
\label{tab:results}
\begin{center}
\begin{tabular}{@{}l|c|c}
\hline
Dataset/ &Normal vs. &Normal vs. \\
Method &Abnormal &Lung opcacities\\
\hline
$\mathbf{U}$ & 0.627$\pm$0.036 & 0.592$\pm$0.021 \\
\hline
$\mathbf{U}+\mathbf{E}$ & 0.687$\pm$0.032 & 0.659$\pm$0.025\\
\hline
$\mathbf{U}+\mathbf{D}$ & 0.737$\pm$0.028 & 0.720$\pm$0.034\\
\hline
$\mathbf{U}+\mathbf{D}+\mathbf{E}$ & 0.841$\pm$0.030 & 0.802$\pm$0.033\\
\hline 
\end{tabular}
\end{center}
\vspace{-2mm}
\end{table}

\begin{figure*}[ht]
  \centering
  \vspace{-2mm}
  \includegraphics[width=0.9\linewidth]{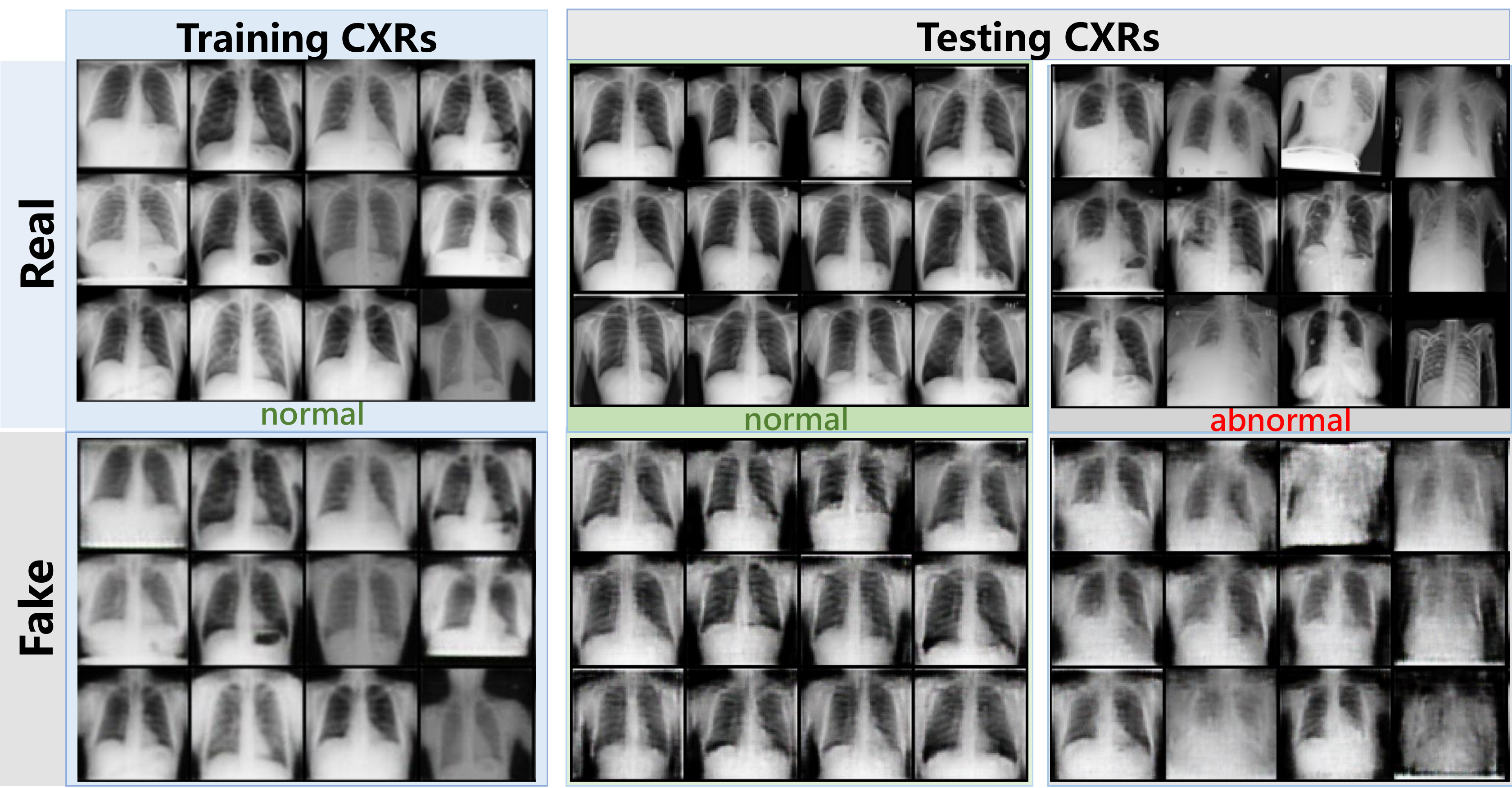}
  \vspace{-2mm}
  \caption{Examples of ground truth CXRs and their corresponding reconstructed images in training and testing stages.}
  \vspace{-2mm}
  \label{fig:results}
\end{figure*}

We show qualitative examples of real CXRs and reconstructed images by our framework in Figure \ref{fig:results}. As can be seen from the figure, our model is able to generate normal CXRs of high quality in the training stage (see left column of Figure \ref{fig:results}). In the testing stage, the proposed model reconstructs normal CXR images (middle column) with much better quality than abnormal CXRs (right column), where not only the abnormal image contents are blurry and messy, but also the geometrical structures of the chest regions are distorted. This confirms how our one-class classifier can indeed differentiate the normal CXRs and abnormal ones. 

\vspace{-1mm}
\section{Conclusion}
\vspace{-1mm}
In this paper, we present an end-to-end trained generative adversarial one-class classifier for abnormal chest X-ray detection, by learning only from normal CXRs. The proposed method is able to reconstruct the normal CXRs while performing poorly on reconstructing the abnormal ones, since the abnormal CXR image contents are unseen during training. Our method is fast and effective, with less manual annotation effort needed. Quantitative and qualitative experimental results demonstrate encouraging performance, showing a potential of reducing workload for radiologists. The proposed method could possibly be extended and applied to other image modalities in future work.

\vspace{-1mm}

\section{Acknowledgments}
\vspace{-2mm}
This research was supported by the Intramural Research Program of the National Institutes of Health Clinical Center and by the Ping An Technology Co., Ltd. through a Cooperative Research and Development Agreement. The authors thank NVIDIA for GPU donation.
\vspace{-1mm}

\bibliographystyle{IEEEbib}
\bibliography{refs}

\begin{thebibliography}{1}

\bibitem{Wang_CVPR2017}
X.~Wang, Y.~Peng, L.~Lu, Z.~Lu, M.~Bagheri, and R.~M. Summers,
\newblock ``Chestx-ray8: Hospital-scale chest x-ray database and benchmarks on
  weakly-supervised classification and localization of common thorax
  diseases,''
\newblock in {\em CVPR}, 2017.

\bibitem{Yates}
E.J. Yates, L.C. Yates, and H.~Harvey,
\newblock ``Machine learning “red dot”: open-source, cloud, deep
  convolutional neural networks in chest radiograph binary normality
  classification,''
\newblock {\em Clinical Radiology}, vol. 73, no. 9, pp. 827 -- 831, 2018.

\bibitem{Tang_MLMI}
Y.~Tang, X.~Wang, A.~P. Harrison, L.~Lu, J.~Xiao, and R.~M. Summers,
\newblock ``Attention-guided curriculum learning for weakly supervised
  classification and localization of thoracic diseases on chest radiographs,''
\newblock in {\em Machine Learning in Medical Imaging}, 2018.

\bibitem{MOYA1996463}
M.~M. Moya and D.~R. Hush,
\newblock ``Network constraints and multi-objective optimization for one-class
  classification,''
\newblock {\em Neural Networks}, vol. 9, no. 3, pp. 463 -- 474, 1996.

\bibitem{sabokrou2018}
M.~Sabokrou, M.~Khalooei, M.~Fathy, and E.~Adeli,
\newblock ``Adversarially learned one-class classifier for novelty detection,''
\newblock in {\em CVPR}, 2018.

\bibitem{Goodfellow_GAN}
I.~Goodfellow, J.~Pouget-Abadie, M.~Mirza, B.~Xu, D.~Warde-Farley, S.~Ozair,
  A.~Courville, and Y.~Bengio,
\newblock ``Generative adversarial nets,''
\newblock in {\em NIPS}, 2014.

\bibitem{Madani18}
A.~Madani, M.~Moradi, A.~Karargyris, and T.~Syeda-Mahmood,
\newblock ``Semi-supervised learning with generative adversarial networks for
  chest x-ray classification with ability of data domain adaptation,''
\newblock in {\em ISBI}, 2018.

\bibitem{Salehinejad}
H.~Salehinejad et~al.,
\newblock ``Generalization of deep neural networks for chest pathology
  classification in x-rays using generative adversarial networks,''
\newblock in {\em ICASSP}, 2018.

\bibitem{Ronneberger}
O.~Ronneberger, P.Fischer, and T.~Brox,
\newblock ``U-net: Convolutional networks for biomedical image segmentation,''
\newblock in {\em MICCAI}, 2015.

\end{thebibliography}

\end{document}